\documentclass[letterpaper, 10 pt, conference]{ieeeconf}  

\IEEEoverridecommandlockouts                              
\usepackage[utf8]{inputenc}
\usepackage[misc]{ifsym}
\usepackage[pdftex]{graphicx} 
\usepackage{subcaption} 
\usepackage{amsmath} 
\usepackage{hyperref}
\hypersetup{
    colorlinks=true,
    linkcolor=blue,
    filecolor=magenta,      
    urlcolor=black,
}
\title{\LARGE \bf Urban Swarms: A new approach for autonomous waste management}

\author{Antonio Luca Alfeo*$^{1,2,4}$, Eduardo Castell\'{o} Ferrer*$^{1}$, Yago Lizarribar Carrillo$^{1}$,\\ Arnaud Grignard$^{1}$, Luis Alonso Pastor$^{1}$, Dylan T. Sleeper$^{1}$, Mario G. C. A. Cimino$^{2}$,\\  Bruno~Lepri$^{3}$, Gigliola~Vaglini$^{2}$, Kent Larson$^{1}$, Marco Dorigo$^{5}$, Alex `Sandy' Pentland$^{1}$\\ \small{$^1$MIT Media Lab, Massachusetts Institute of Technology, Cambridge, USA}\\
\small{$^2$University of Pisa, Pisa, Italy}
\small{$^3$Bruno Kessler Foundation, Trento, Italy}
\small{$^4$University of Florence, Florence, Italy}\\
\small{$^5$IRIDIA, Universit\'{e} libre de Bruxelles, Brussels, Belgium}\\
$*$These authors contributed equally to this work}

\begin{document}
\maketitle
\thispagestyle{empty}
\pagestyle{empty}
{\let\thefootnote\relax\footnote{{Corresponding author: Eduardo Castell\'{o} Ferrer, ecstll@media.mit.edu}}}
\vspace{-0cm}
\begin{abstract}

Modern cities are growing ecosystems that face new challenges due to the increasing population demands. One of the many problems they face nowadays is waste management, which has become a pressing issue requiring new solutions. Swarm robotics systems have been attracting an increasing amount of attention in the past years and they are expected to become one of the main driving factors for innovation in the field of robotics. The research presented in this paper explores the feasibility of a swarm robotics system in an urban environment. By using bio-inspired foraging methods such as multi-place foraging and stigmergy-based navigation, a swarm of robots is able to improve the efficiency and autonomy of the urban waste management system in a realistic scenario. To achieve this, a diverse set of simulation experiments was conducted using real-world GIS data and implementing different garbage collection scenarios driven by robot swarms. Results presented in this research show that the proposed system outperforms current approaches. Moreover, results not only show the efficiency of our solution, but also give insights about how to design and customize these systems.  

\end{abstract}

\section{Introduction}
\label{introduction}

Swarm robotics systems \cite{DorBirBra2014:sch-sr} have the potential to shape the future of many applications, e.g. targeted material delivery \cite{Andrea2008}, precision farming \cite{Bechar2017a} and search task \cite{cimino2016using}. Assisted by technological advancements such as distributed computing \cite{Ferrer2016c}, novel hardware design \cite{Mondada2006}, and manufacturing techniques \cite{Oxman2014,CastelloFerrer2015}, nowadays swarms of robots are starting to become an important part of industrial activities such as warehouse logistics \cite{Andrea2012,Liu2017}. The potential of robot swarms has been acknowledged as one of the ten robotics grand challenges for the next 5-10 years that will have notable socioeconomic impact \cite{Yang2018a}. 
Currently, one of the main study areas of swarm robotics systems is on foraging behaviors. Foraging is the set of actions to explore and collect objects or information scattered in an environment. Foraging tasks can be projected to more complicated problems (e.g., exploration vs exploitation trade-offs, consumer and producer models, etc.), and currently they are one of the main benchmarks to evaluate swarm robotics systems \cite{Lu2018b}. Applications of swarm robotics foraging are wide-ranging from carrying objects and tokens to specific target locations \cite{Dorigo2005, Castello2016} to rescuing natural disaster victims \cite{Payton2005}. The similarity among these examples is that robots always leave from and return to a common central location (e.g., nest, headquarters, etc.). Central Place Foraging (CPF), as it is called, is in fact the most studied foraging approach in the field \cite{Winfield2009,Brambilla2013,Castello2016}. Although CPF provides good results in simple missions and indoor scenarios, the overall performance (e.g., tokens collected, packages delivered, etc.) of the swarm decreases when the size of the scenario or the number of robots grow \cite{Lu2016,Zia2017b}. Due to this phenomenon, CPF-based systems might be inadequate for deployment in larger, more dynamic areas such as big cities or vast urban environments \cite{Salvini2018g}.
However, one possible solution to this issue could be Multiple Place Foraging (MPF). MPF is a bio-inspired problem \cite{Chapman1989a,Schmolke2009a} that relies on multiple nests rather than one central depot. Nests are scattered across the area and each robot inside the swarm can change its correspondent nest depending on its location and energy status \cite{Lu2016a,Lu2018b}. One of the main coordination mechanisms to steer the swarm is stigmergy \cite{zedadra2015design}. With stigmergy, pheromones are released in a shared environment and are used as a type of ``in-field'' communication that can be used to self-organize the swarm collective motion \cite{alfeo2018swarm}. Theoretically, CPF and MPF have a very similar set of parameters \cite{Lu2016a, Lu2016}. However, MPF-based research has not been conducted outside simplistic scenarios \cite{Halasz2007a, Berman2008} and therefore further studies are required to test its feasibility.  
In the meantime, the world is urbanizing at an unprecedented rate~\cite{Shahrokni2014b}. UN-Habitat estimates that by 2050, 3.5 out of the 9.1-billion global residents will be living in informal urban communities \cite{habitat2016urbanization}. On the one hand, novel urban infrastructures together with new technologies such as IoT, 5G, LiDAR, etc. allow to understand the city as a senseable, programmable, and actuable ecosystem \cite{Shahrokni2014b}. On the other hand, this urbanization implies important social and environmental challenges such as fuel production, air pollution, etc. \cite{Girardet2008b, Kumar2015b}. Experts estimate that it will require 57 trillion USD to adapt traditional heavy infrastructures to the informal urban needs \cite{jahan2017human} and that today's solutions will not be able to scale at the pace urbanization is taking place. One of the main urban services that could drastically benefit from the inclusion of novel technology is waste management due to its economical and environmental impact~\cite{Zanella}. For instance, in areas that are experiencing fast growth, waste management has become a challenge since the basic resources are not adapted to such changes \cite{Offenhuber2012}.

The aim of this paper is to explore the synergy of swarm robotics systems and urban environments by using MPF and stigmergy to improve the efficiency and autonomy of the urban waste management system. To achieve this, a diverse set of simulation experiments was conducted using real-world GIS data and implementing different garbage collection scenarios driven by robot swarms. 


\section{Approach Description}
\label{sec:description}

\begin{figure}[tbh]
\centering
  \includegraphics[width=0.92\linewidth]{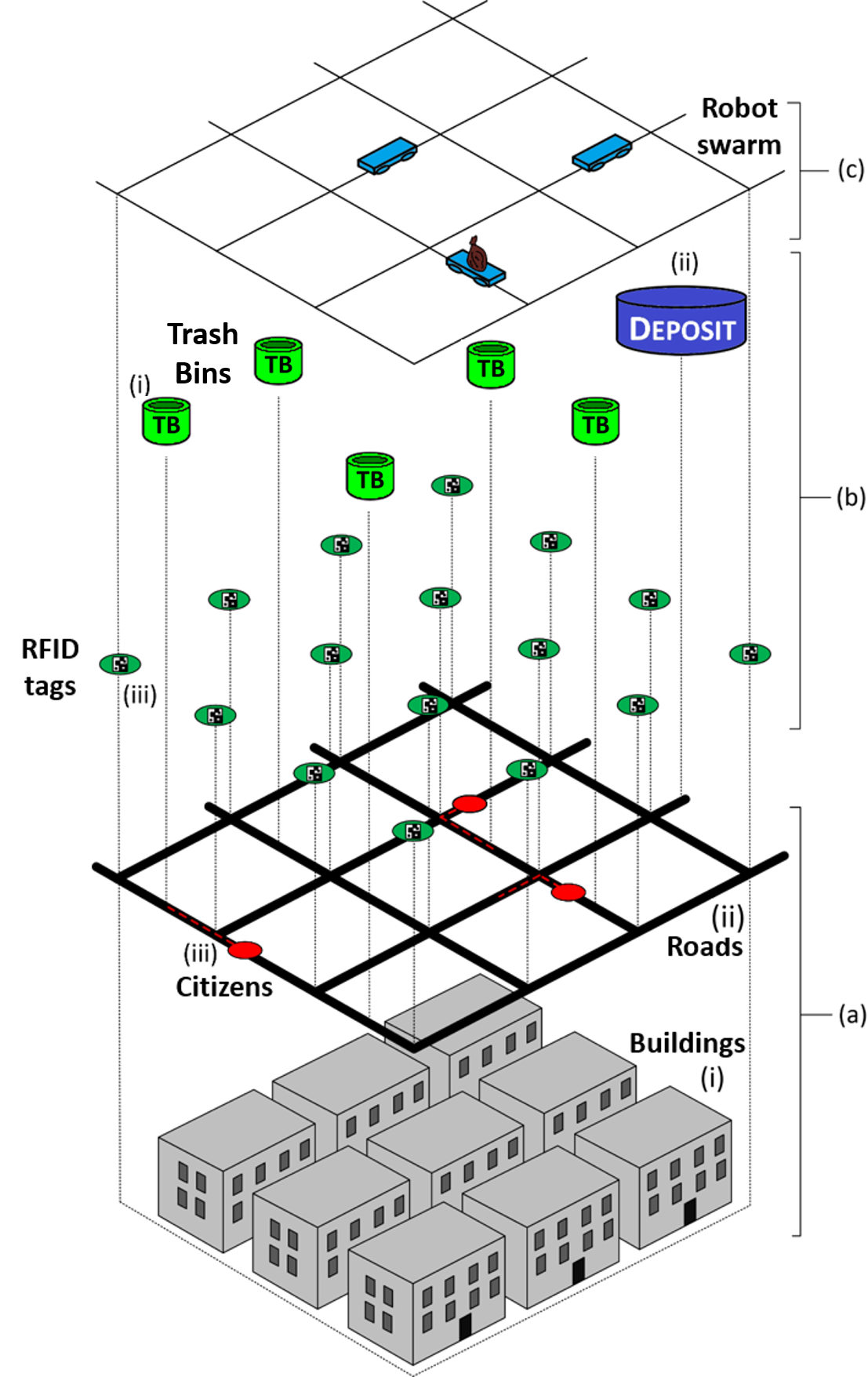}
  \caption{Multilayer model of the ``smart'' city. (a) the urban environment, where buildings, roads, and citizens coexist. (b) the waste management infrastructure, where RFID tags coexist with Trash Bins (TB) and Deposits (Ds). (c) the actuation layer, where swarm robots exploit other layers to deliver the urban waste management service.}
    \label{img:city_layers}
\end{figure}

Nowadays, cities have to respond to the growing demands of more efficient, sustainable, and increased quality of life, thus making them ``smarter''. In this context, the ``smartness'' can be defined as the capability to gain insights about the current urban conditions, and to react dynamically to manage them properly \cite{goodspeed2014smart}. According to this view, smart cities can be seen as cybernetic urban environments where different agents (e.g., citizens) and actuators (e.g., swarm of robots) exploit the city wide infrastructure as a medium to operate synergistically. The approach proposed in this work is presented as a multilayer simulation model where each layer represents one of these components: (a) The urban environment, (b) the waste management infrastructure, and (c) the actuation layer (see Fig. \ref{img:city_layers}).

\subsection{The urban environment}
\label{sec:UrbanEnvironmentLayer}

 The urban environment (Fig. \ref{img:city_layers} (a)) is modelled according to publicly available demographic data and considering a trade-off between the proper representativeness and the complexity of the scenario, thus containing (i) \textit{buildings}, where agents stay in specific hours of the day, (ii) \textit{roads}, used by agents to move between buildings, and \textit{citizens} (iii). Citizens are special agents that move between \textit{buildings} (e.g., home, workplace, amenities, etc.) at certain hours during the day using \textit{roads}. In our approach, citizens recreate the daily activity of the urban area; their simulated behavior and mobility patterns were described in recent literature \cite{grignard2018impact, Alonso2018}.

\begin{figure}[tbh]
  \centering
  \includegraphics[width=0.8\linewidth]{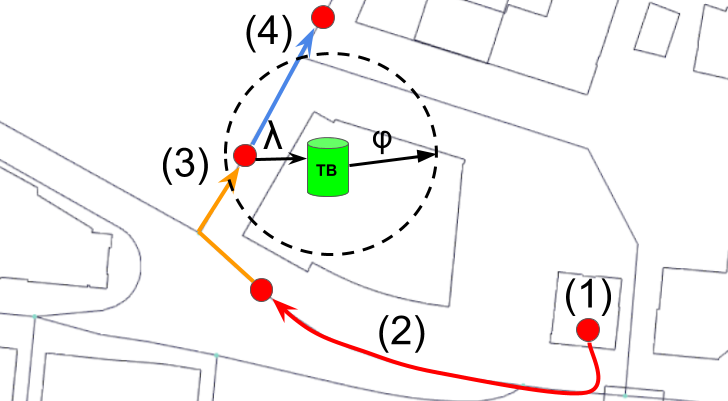}
  \caption{Representation of a citizen's travel path and waste depositing process. Citizens start their activity in an initial location (1). By travelling around (2) citizens generate waste. If citizens find a Trash Bin (TB) within distance $\varphi$ (3), they drop $\lambda$ amount of waste inside. Once this process is completed, or if the TB is full, they continue their journey (4).}
  \label{fig:citizens_fsm}
\end{figure}

Due to the citizens' activity (e.g., shopping, eating out, etc.), waste is generated and deposited in urban Trash Bins (TB). Waste generation is a multi-step process (Fig. \ref{fig:citizens_fsm}). Firstly, citizens are positioned on an initial location (1). When it is time to travel (2) (e.g., go to work, return home, etc.), the citizen chooses a destination and starts the trip. While travelling, if the citizen is bringing waste, is within a distance $\varphi$ from a TB, and the TB is not full, the citizen drops $\lambda$ liters of waste in the TB (3). After depositing waste in the TB, the citizen continues traveling (4). In case citizens find a TB that is already full, they do not drop any waste and continue traveling. 

\subsection{The waste management infrastructure}

\begin{figure}[tbh]
\centering
  \includegraphics[width=0.75\linewidth]{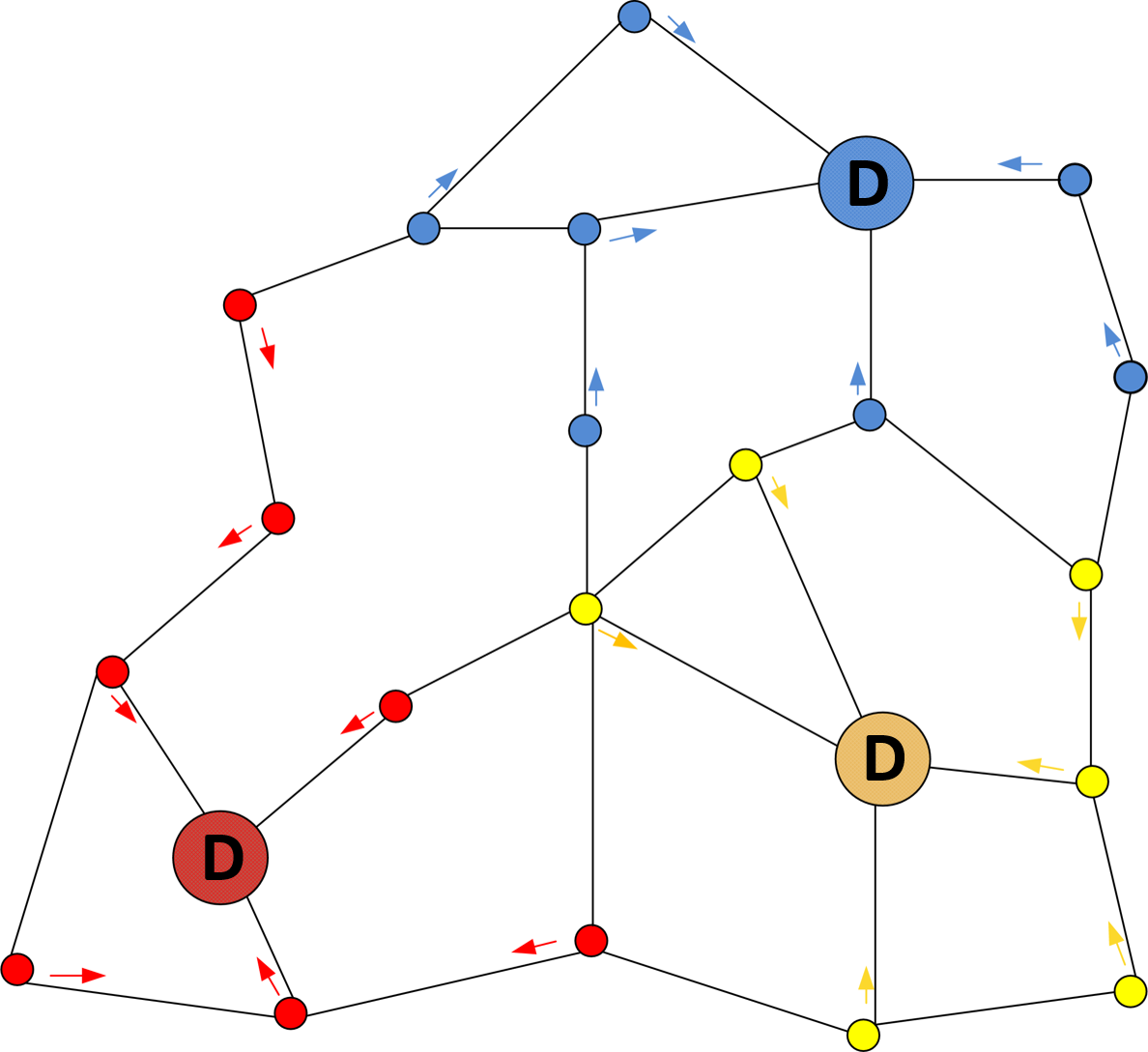}
  \caption{Representation of a scenario with 3 deposits (Ds) and the direction between each road crossing to its closest D.}
    \label{img:directions}
\end{figure}


 On top of the urban environment layer we have the waste management infrastructure (Fig. \ref{img:city_layers} (b)), which employs (i) Trash Bins (TBs). TBs are geolocated and arranged beside the roads\footnote{More information about the geolocation of TBs for the area of study can be found in Sec. \ref{sec:experiment}}. Each TB has an RFID tag containing a unique ID and the current amount of waste inside it \cite{ghadage2017iot}. The TB detects its amount of waste and updates the RFID tag accordingly. Once a minimum threshold is exceeded, the TB automatically packs the waste into a transportable unit~\cite{chomik2017waste}. The number of packed waste units that can be kept in each TB is limited. Once this limit is exceeded, the TB is no longer usable. (ii) Deposits (Ds) are facilities that provide final trash disposal services (e.g., waste compactors, recycle processes, etc.) as well as robot battery refills. (iii)~RFID tags at every crossroad store the information needed to steer the swarm of robots. Specifically, each RFID tag contains the time-stamp of the last RFID operation, the amount of pheromone characterizing each road on that crossroad, and the distance and the direction toward the closest D. The path and distances between each road crossing and the closest D are fixed and known. This information allows robots to compute the shortest path between each crossing and the closest D and to store this information in its correspondent RFID tag. As an example, in Fig. \ref{img:directions} a scenario with 3 Ds is shown. 
 
\subsection{The actuation layer}
\label{sec:actuationlayer}

\begin{figure}[tbh]
	\centering
	\includegraphics[width=\linewidth]{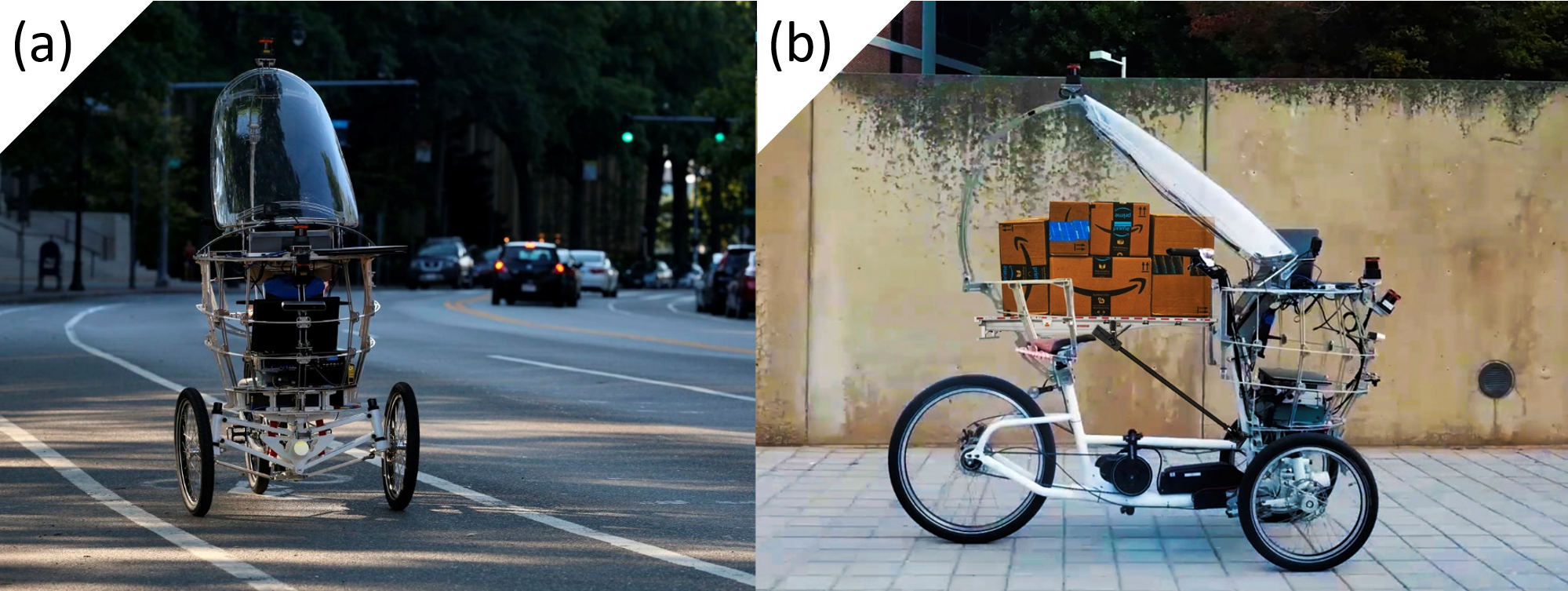}
    \caption{(a) The PEV on the streets of the Kendall area (Cambridge, MA). (b) The PEV carrying several packages as a payload.}
	\label{fig:PEV}
\end{figure}

The actuation layer (Fig. \ref{img:city_layers} (c)) is composed of a swarm of robots in charge of carrying the waste from each TB to the closest D. In order to increase the feasibility of our approach, we decided to model our robots using a real-world platform with specifications suited for the task. The Persuasive Electric Vehicle (PEV), depicted in Fig. \ref{fig:PEV}, is an autonomous tricycle developed at the MIT Media Lab aimed to be a hybrid between autonomous cars and bike sharing systems. 

The core idea behind the PEV is to provide an affordable, highly-customizable, self-driving solution to urban mobility. One of the main advantages of this platform over more ``traditional'' approaches is that it can operate on bike lanes; therefore, it would not stress the already saturated road infrastructure of a populated urban area. The main specifications for the PEV are: a maximum payload of 120.0 kg, maximum speed of 40.0 km/h, and 2 hours battery autonomy. In addition, the PEV is equipped with a wide variety of sensors such as R/W RFIDs, cameras, LIDARs, IMUs, etc. 

\begin{figure}[tbh]
\centering
  \includegraphics[width=0.95\linewidth]{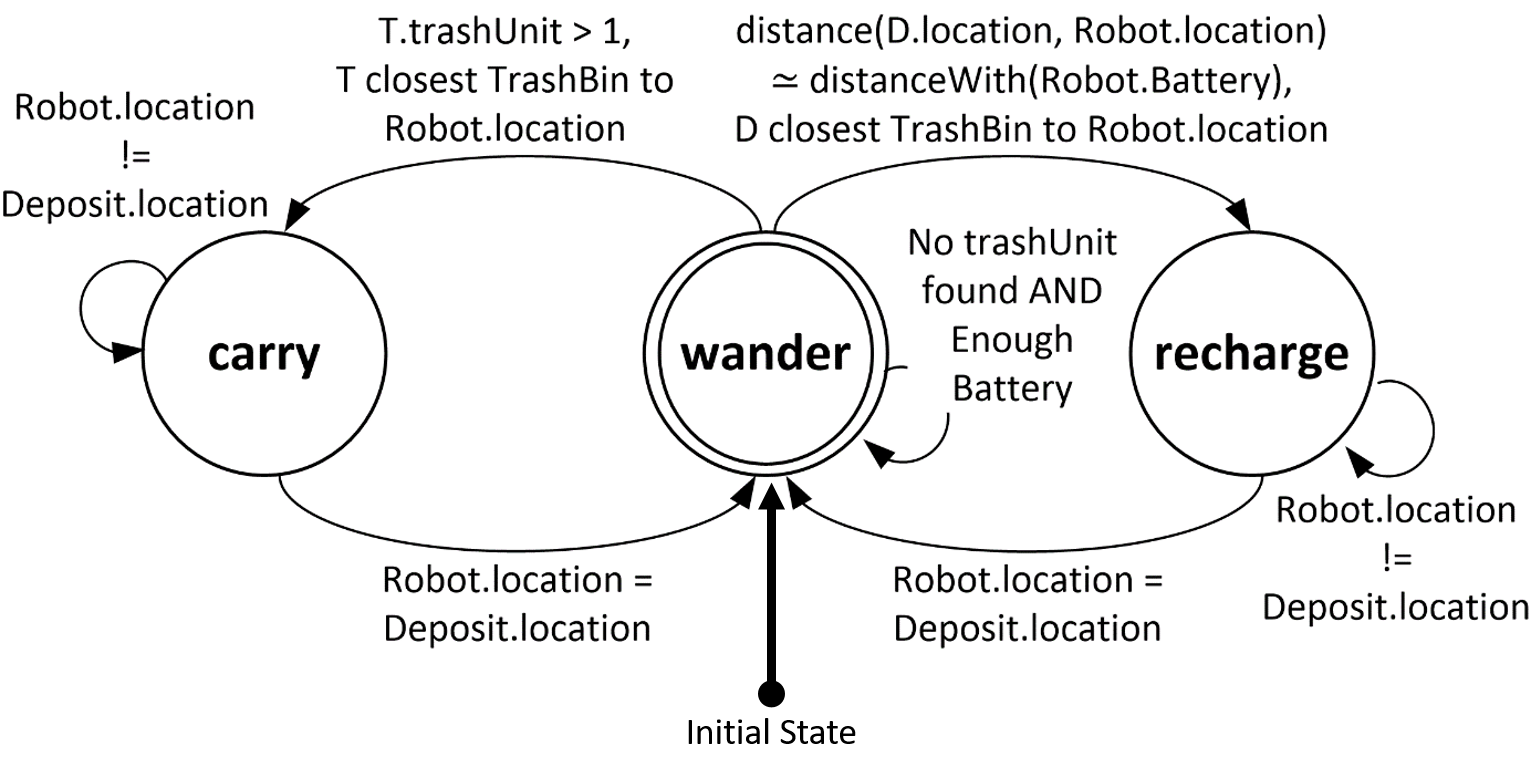}
  \caption{Robots' behavior as a finite state machine: robots \textbf{Wander} to explore the environment. Robots move to the closest D to \textbf{Carry} the waste they find in a TB, or to \textbf{Recharge} their battery before it gets exhausted.}
  \label{img:FSM_ROBOT}
\end{figure}
Fig. \ref{img:FSM_ROBOT} depicts the robot behavior via a finite state machine diagram with 3 states. The initial state of every robot is \textbf{Wander}. While in the wandering state, the robot travels from one road crossing to another by choosing the road with the strongest pheromone level. If this is not possible (i.e., if there is no pheromone on any road or all the roads have the same amount of pheromone), the next edge is chosen randomly. In order to avoid getting stuck on the same area until all the pheromones evaporate, robots can choose a random road with a probability of (1-$X_{r}$), where $X_{r}$ represents the exploitation rate of the robots, a parameter that will be described in Sec. \ref{sec:experiment}.

At each road crossing, the robot estimates the distance that can be traveled with the remaining battery. If this distance is approximately the same to the closest D, the robot state switches to \textbf{Recharge}. In this state, the robot moves towards the closest D by following the directions on the RFID tags on the way. The robot returns to the wander state when it reaches the closest D, since we assume that the deposit automatically swaps the battery of the robot.

When the robot is close to a TB, it reads the RFID tag on the TB. If the TB contains at least one transportable waste unit, the robot changes its state to \textbf{Carry}. Then, the robot withdraws a waste unit from the TB and moves to the closest D by following the directions available on the RFID tags on the way. While the robot is in the carry state it ignores the TBs in its way. The robot returns to the wander state after reaching the D.

The fundamental mechanism on which the self-organization of the swarm of robots is based is the perception and distribution of pheromone amounts. In this work, an RFID tag is placed at each road crossing. In addition to the direction to the closest D and the time-stamp of the last operation, the RFID tag contains the pheromone amounts corresponding to each direction (thus, each road) that can be taken from the road crossing. The pheromones amounts are maintained in a consistent state by the robots, which manipulate them according to a precise set of rules that echo biological models of stigmergy-based foraging~\cite{zedadra2017multi}.
In particular, the pheromone amount is subject to three processes:

\textbf{Marking}, i.e., the addition and aggregation of pheromone to the already existing pheromone trail due to the performance of a given action (e.g., when an ant is carrying food). In our model, this is achieved by robots in the carry state by increasing the amount of pheromone in the crossroads from which the robot is coming, thus marking the path towards where the waste is being generated. This amount is proportional to the amount of waste found in the TB.

\textbf{Evaporation}, i.e., the decay of the pheromone trail over time. In our model, this is achieved by each robot by decreasing the amount of pheromone corresponding to the current crossroad visited. The amount of pheromone decreasing through the evaporation mechanism is proportional to the difference between the current time instant ($t$) and the time-stamp ($ts$) of the last RFID operation. If the final amount of pheromone is less than zero, it is set to zero.

\textbf{Diffusion}. To increase the probability that robots are attracted to a location with uncollected waste, we implemented the diffusion mechanism: the capability of marking a road with a small portion of the pheromone perceived on the last RFID tag, so as to make the pheromone perceptible even from roads immediately close to the marked path and steer the robots toward it. In our model, this is achieved by each robot by increasing the amount of pheromone corresponding to the road from which the robot is coming.

In brief, when the robot interacts with an RFID tag, it decreases the amounts of pheromones on it depending on the time elapsed since the last RFID operation and the evaporation rate. Moreover, the amounts of pheromones regarding the direction (i.e., the road) from which the robot comes from is increased due to the diffusion and marking processes (if the robot is carrying waste). Specifically, the following formula describes the updating procedure of the amount of pheromone corresponding to each edge in the RFID tag: 

\small
\begin{equation}
  \vspace*{0.8cm}
\label{eq:pheromoneUpdate}
P_{t} =  P_{ts}-[E_{r} \cdot P_{a} \cdot (t-ts)]+(P_{a} \cdot T_{a})+(D_{r} \cdot P_{max})
  \vspace{-0.8cm}
\end{equation}
\normalsize

In Eq. \ref{eq:pheromoneUpdate}, $P_{t}$ represents the amount of pheromone corresponding to the current edge at the current time instant. $P_{ts}$ is the amount of pheromone corresponding to the current edge at the time-stamp (i.e., the last operation on the RFID tag). $E_{r}$ is the evaporation rate (0 $\leq$ $E_{r}$ $\leq$ 1), i.e., the amount of the pheromones disappearing per unit of time. $P_{a}$ is the amount of pheromone to be added to the RFID tag for each unit of waste found in the TB from which the waste has been picked up (only if the robot is performing the carry action). $T_{a}$ is the amount of waste found in the TB (only if the robot is performing the carry action and it comes from the current edge). $D_{r}$ is the diffusion rate (0 $\leq$ $D_{r}$ $\leq$ 1), in other words, the portion of the pheromone to diffuse. Finally, $P_{max}$ is the maximum pheromone amount on the last RFID tag.  


\section{Experimental Setup}
\label{sec:experiment}

The presented system\footnote{A copy of code repository can be found here: https://goo.gl/tqRvS4} was developed in GAMA \cite{grignard2013gama, taillandier2018building}; a realistic agent-based simulation tool applied in fields such as urban planning, disaster mitigation, etc. The urban environment layer described in Sec. \ref{sec:UrbanEnvironmentLayer} was built using real-world GIS data by integrating the map of the Kendall (Cambridge, MA) urban area. The number of citizens was initialized to 10,000 following previous research works about the area of study \cite{grignard2018impact,Alonso2018}. Regarding the waste generation process, according to the EPA (Environmental Protection Agency) Americans produce 2 kg of waste per day \cite{EPA2012}. However, not all of that waste goes into public TBs; a large portion of it is dropped in residential bins as well. According to \cite{MasDEP2013}, Cambridge public works collect an average of 1.18 kg of waste per citizen per day from Cambridge households. Thus, we estimated that 0.82 kg of waste per citizen was deposited in public TBs everyday. By using conversion data about the weight of different types of waste \cite{ConversionFactorsEPA2016}, we transformed the amount of kg of waste generated per citizen into liters. The result of these conversions was 8.42 liters/citizen. We initialized $\lambda$ to this value throughout our simulations. Finally, $\varphi$ was initialized to 50 meters.

\begin{figure}[tbh]
\centering
  \includegraphics[width=\linewidth]{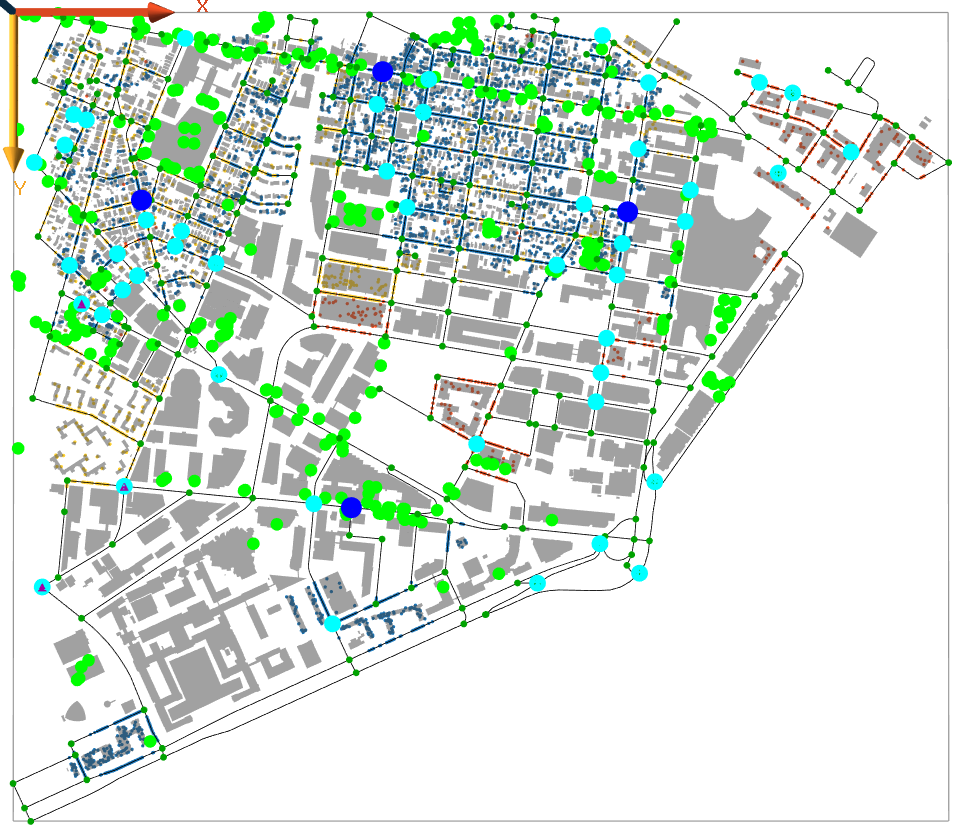}
  \caption{Urban swarm's simulation, Kendall area (Cambridge, MA). Citizens (small colored dots) can drop waste in close TBs (green dots) when moving on the road. The robots (cyan dots) can move over the roads to carry the waste from the TBs into the Ds (blue dots) using the information in the RFID tags at each crossroad.}
    \label{img:simulation}
\end{figure}

Fig. \ref{img:simulation} shows a screenshot of the simulation tool where each building (gray), TB (green dots), road (black lines), and citizen (small dots over buildings and roads) is depicted. Moreover, we obtained the number and location of TBs from the Cambridge Geographic Information Service\footnote{More information can be found here: https://goo.gl/os3nxN}. Ds (dark blue dots) were placed by using the k-means algorithm to minimize the distance between them and the TBs. 

Finally, each robot is represented as a cyan dot. The effectiveness of the proposed approach was tested with different configurations (i.e., by changing the behavioral and scenario parameters). The ranges of the parameters were chosen to allow the implementation of significantly different behaviors on the swarm of robots. Specifically: 

\textbf{Number of Robots} ($R_{n}$): affects the effectiveness and the size of the overall system; the tested values are 20, 35, and 50 robots.

\textbf{Evaporation Rate} ($E_{r}$): affects the amount of time the system retains the information about the waste disposal demand; the tested values are 0.05$\%$, 0.15$\%$, and 0.3$\%$.

\textbf{Exploitation Rate} ($X_{r}$): affects the probability that the robots follow the path with the strongest pheromone rather than a random one. A higher value results in a higher exploitation of the information about the waste disposal process, whereas lower values increase the exploration of the overall scenario. A more exploratory swarm easily reaches isolated TBs, while a swarm more prone to the exploitation of the waste disposal information exhibits a more aggressive waste collection behavior toward the previously-discovered non-empty TBs. The exploitation rate affects also the diffusion rate, which is 1-$X_{r}$, since in a non-exploratory swarm the diffusion will just reinforce the already marked path. The tested values are 0.6, 0.75, and 0.9.

\textbf{Carriable Waste} ($C_{w}$): affects the amount of carriable waste per robot. Lower values of this parameter result in a more responsive but slow reaction of the system since the waste can be picked only if the $C_{w}$ is already present in the bin. Indeed, we assume that the $C_{w}$ corresponds to the amount of waste that a TB can pack to be transported. The tested values were 6, 12, and 18 liters of waste. This was designed taking into account the PEV capabilities introduced in Sec. \ref{sec:actuationlayer}.

\textbf{Number of Deposit} ($D_{n}$): affects the responsiveness of the overall system; the tested values are 2, 3, and 5 Ds. 

\subsection{Current waste management model}
In order to provide a better insight about the implications of the proposed approach, we decided to compare it with the waste management model that is currently operating in the urban area of study (i.e., truck-based). We built this model based on the information provided by the Cambridge Department of Public Works (CDPW)\footnote{More information about the specific route and timetables can be found here: https://goo.gl/cHXDYS}. According to CDPW, a single truck system in 5 working days (Monday-Friday) in 5 hours a day (7-12PM) is able to empty approximately 6000 TBs. This results in a capability of emptying about 240 trash bins in an hour. In our scenario, the number of TBs is fixed to 274. Therefore, the truck should be able to empty all TBs in about an hour and 10 minutes and should pass once a day. 

\section{Results}
\label{sec:results}
We conducted ten simulations for each possible parameter combination introduced in Sec. \ref{sec:experiment}. In order to analyze the effectiveness of both approaches, we introduced two performance metrics. First, the \textbf{Amount of Uncollected Trash} (AUT, measured in liters) represents the amount of waste left unattended in the environment. Higher AUT levels correlate to the appearance of urban issues such as pests, air pollution, and public health problems. Second, the average number of \textbf{Full Trash Bins} (FTB, measured in units) in the scenario during a day. FTB shows the average number of unusable TBs that the system leaves in the urban environment during the day. Higher FTB values typically correlate to higher citizens' dissatisfaction rates since they might have to travel longer distances to dispose their waste. For the sake of interpretability, each of the measures is presented as a percentage (the lower, the better) considering that in our scenario there are 274 TBs with a capacity of 125 liters each\footnote{This capacity correlates to the TB model (Big Belly BB5) currently operating in the study area. The BB5 is equipped with a solar-powered waste compactor and a wireless data link. More information can be found here: http://bigbelly.com/}.
 
\begin{figure}[tbh]
\centering
  \includegraphics[width=\linewidth]{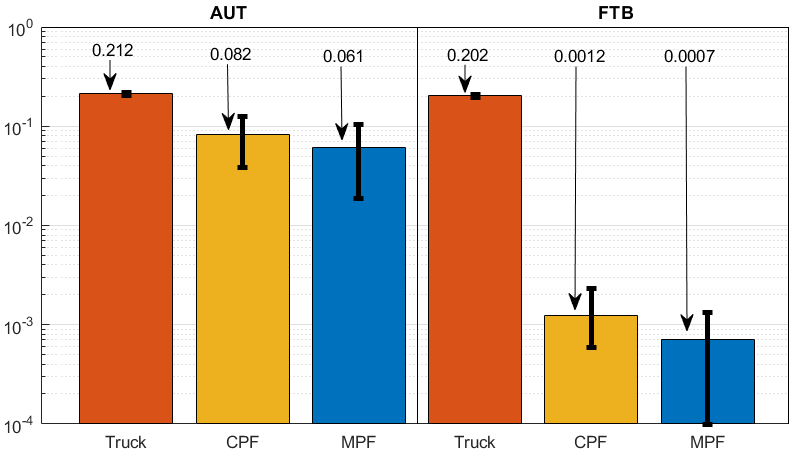}
  \caption{Percentage of AUT and FTB measures for the truck and the swarm (best parameterizations) with the CPF and the MPF approach. Log scale.}
  \label{img:TruckVsSwarm}
\end{figure}

\begin{figure*}[t!]
  \begin{center}
  \begin{subfigure}[t]{0.45\textwidth}
    \includegraphics[width=\textwidth]{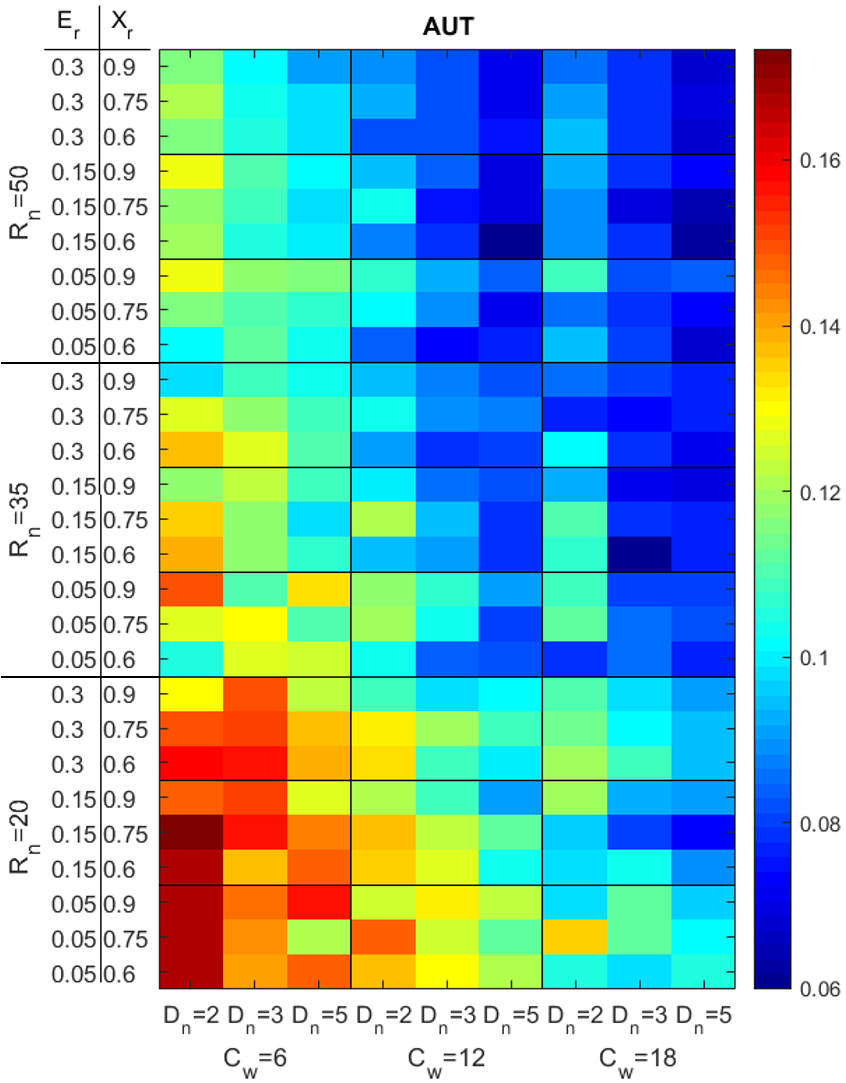}
    \caption{}
    \label{img:allResultsTrashAmount}
  \end{subfigure}
  \hspace{1.5cm}
  \begin{subfigure}[t]{0.45\textwidth}
    \includegraphics[width=\textwidth]{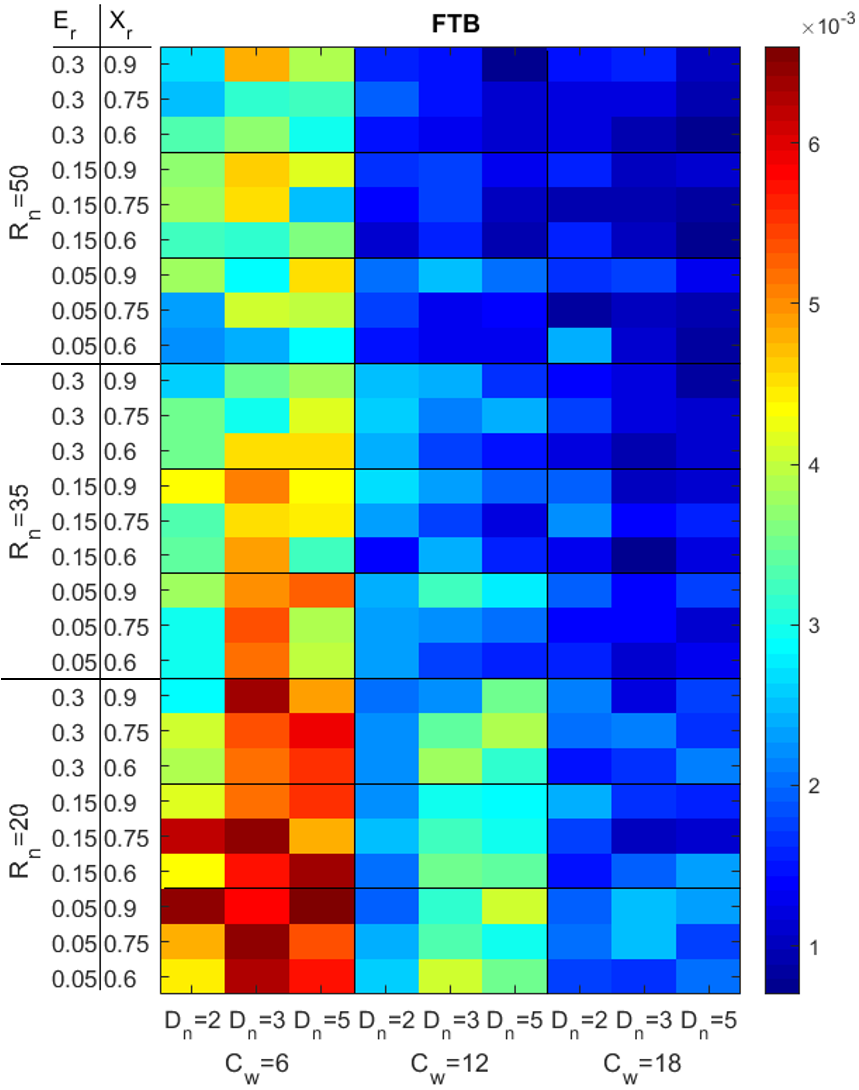}
    \caption{}
    \label{img:allResultsFullBins}
  \end{subfigure}
  \caption{(a) The average percentage of Amount of Uncollected Trash (AUT) in the scenario during a day by changing the parameterization. (b) The average percentage of Full Trash Bins (FTB) in the scenario during a day by changing the parameterization.}
  \end{center}
  \vspace{-0.5cm}
\end{figure*}

We compare the performance of our proposed approach against (i) the current trash disposal model i.e., truck-based model; and (ii) a CPF solution i.e., by using the stigmergy-based foraging with a single deposit. Thus, we simulated each model and computed the corresponding performance metrics. The results obtained with ten simulations are summarized in Fig.\ref{img:TruckVsSwarm}. Results show that the current waste management system offers lower performance than the swarm-based solution proposed in this paper. In fact, in terms of the percentage of AUT and FTB, the MPF approach offers a decrease of 71\% (0.061) and 99\% (0.0007) compared to the results obtained with the truck-based model (0.212 and 0.202 respectively). Moreover, the MPF approach results to be more effective than the CPF approach for both AUT and FTB. 

It can be observed from the results in Fig. \ref{img:allResultsTrashAmount} that our system is able to collect most of the disposable trash on average during the day, leaving a AUT of 17\% (worst case) and 6\% (best case). Moreover, in general, the increase of $D_{n}$, $C_{w}$, and $R_{n}$ results in a lower AUT. A greater number of deposits results in a shorter path to reach them, while more robots and a greater carriage capacity result in a system that collects and disposes waste more quickly. 

The results in Fig. \ref{img:allResultsFullBins} prove that our system is able to empty the TBs fast enough to have less than 1 (0.0007\%, best case) or 2 (0.0066\%, worst case) FTB in the scenario on average during the day. Moreover, it can be noticed that by increasing $E_{r}$ and $R_{n}$ we obtain a lower percentage of FTB since it increases the responsiveness of the system. 

\section{Discussion}
\label{sec:discussion}

\begin{table}[tbh]
 \begin{center} 
\begin{tabular}{|c|c|c|c|}
\hline
                 & $\beta_{Rn}$ & $\beta_{Cw}$ &  $\beta_{Dn}$ \\
\hline
\textbf{AUT}    & -0.6615   &  -0.7580 & -0.3868 \\
\textbf{FTB}    &  -0.4647  &  -0.9584 &  0.0407 \\
\hline
\end{tabular}
\caption{$\beta$ coefficients, multiple standardized regression}
\vspace{-0.5cm}
\label{table:stdregression}
\end{center}
\end{table}

Two parameter configurations provided the best performances: (1) $R_{n}$=35, $E_{r}$=0.15, $X_{r}$=0.6, $C_{w}$=18, $D_{n}$=3 which produces a 0.7\% FTB and 6.1\% AUT and (2) $R_{n}$=50, $E_{r}$=0.15, $X_{r}$=0.6, $C_{w}$=12, $D_{n}$=5 which obtains a 0.9\% FTB and 5.9\% AUT. The first solution is characterized by a medium number of deposits and robots, but can assure a good responsiveness of the system thanks to the relatively high $C_{w}$. On the other hand, the second solution, is characterized by a large $D_{n}$ and $R_{n}$, and a medium $C_{w}$. These configurations suggest that there is a balance between the size of the system ($D_{n}$ and $R_{n}$) and the amount of carriable waste ($C_{w}$). At the same time, both solutions are characterized by an $E_{r}$ of 0.15 and a relatively high $X_r$ with a value of 0.6. 
To assess the relationship between $R_{n}$/$D_{n}$/$C_{w}$ and the proposed performance measures, we computed a multiple standardized regression, i.e., the regression in which both dependent and independent variable are substituted by their Z-score \cite{keith2014multiple}.
By considering the size of each $\beta$ coefficient, we are able to compare the impact of each variable on the corresponding performance measure despite the differences in their scale. Larger coefficients correspond to higher contributions, whereas the sign describes the direction (positive or negative) of the contribution. It is worth recalling that all performance measures proposed must be minimized to achieve a better performance. Table \ref{table:stdregression} shows that increasing $R_n$ and $C_w$ improves the performance. Surprisingly, increasing $D_n$ increases the AUT performance but decreases FTB's; indeed, a greater number of Ds means shorter travel distances for robots, therefore a more responsive system and a better AUT performance. However, since Ds are the destination of all robots, the paths around the D are more likely to be marked by digital pheromones which are aimed at steering the swarm. This means that around each D the robots' exploratory capability is reduced due to the overwhelming amount of deployed pheromones, therefore the probability of reaching TBs in areas that require to pass by a near D may be lower. This explains the decrease in the performance at the bottom-left corner of Fig. \ref{img:allResultsFullBins} when $D_{n}$ is increased.

\section{Conclusions}
\label{conclusions}
In this paper, the design, implementation, and experimentation of a robotic swarm aimed at managing the waste disposal in a realistic urban environment was presented. We showed that a swarm is able to handle the waste management in an effective and self-organized manner, without any external information source or prior knowledge about the trash disposal demand. With the proposed approach, both the average amount of trash and the average number of full trash bins during the day are considerably reduced compared to the current solution. Moreover, we provided insights on how to parameterize the system according to the desired outcome, that is, higher exploration (needed to reduce the FTB) or more exploitation (needed to reduce the AUT). Finally, the proposed approach is not specific to waste management and can be used in a number of different applications such as package delivery, autonomous vehicle rides, etc. To prove the suitability of the proposed approach with different applications future research work will be focused on real world contexts and scenarios.

\section*{Acknowledgment}
\label{acknowledgment}
This project has received funding from the European Union’s Horizon 2020 research and innovation programme under the Marie Skłodowska-Curie grant agreement No. 751615.

\bibliographystyle{ieeetr}
\bibliography{References}

\end{document}